\documentclass[conference]{IEEEtran}
\IEEEoverridecommandlockouts
\usepackage{cite}
\usepackage{amsmath,amssymb,amsfonts}
\usepackage{algorithmic}
\usepackage{graphicx}
\usepackage{textcomp}
\usepackage{xcolor}
\def\BibTeX{{\rm B\kern-.05em{\sc i\kern-.025em b}\kern-.08em
    T\kern-.1667em\lower.7ex\hbox{E}\kern-.125emX}}
\begin{document}

\title{Flower Across Time and Media:  Sentiment Analysis of Tang-Song Poetry and Visual Correspondence\\

}
\author{
\IEEEauthorblockN{Shuai Gong*}
\IEEEauthorblockA{\textit{Faculty of Humanities and Arts} \\
\textit{Macau University of Science and Technology} \\
Macao, Macao SAR, China \\
shinegs@qq.com}

\and

\IEEEauthorblockN{Tiange Zhou*\thanks{*These authors contributed equally to this work. Gong has been working on the data collection and selection while Zhou has been working on the computational analysis and data visualization.}†\thanks{†Corresponding author}}
\IEEEauthorblockA{\textit{School of Future Design} \\
\textit{Beijing Normal University} \\
Zhuhai, China \\
tiangezhoumusic@gmail.com}
}
\maketitle

\begin{abstract}
This study explores the dynamic interplay between
literary sentiment and visual culture in Tang-Song China by
employing BERT-based sentiment analysis to track evolving
emotional connotations of predominant floral motifs in classical
poetry. Focusing on peony in Tang and plum in Song blossom
imagery, the research identifies sentiment shifts across the Tang
(618–907) and Song (960–1279) dynasties and correlates these
changes with contemporaneous developments in paintings. The
findings demonstrate that poetic sentiment not only blects but
also anticipates broader aesthetic trends in material culture,
establishing a data-driven framework for cross-media analysis
of symbolic evolution. By integrating computational linguistics
with cultural historiography, this work advances AI-assisted
methods for decoding the bidirectional influence between textual
expression and visual design in Tang-Song China societies.
\end{abstract}

\begin{IEEEkeywords}
Computational Cultural Heritage Analysis, BERT-enabled Symbolism Decoding, Aesthetic Intelligence
\end{IEEEkeywords}

\section{Introduction}

The Tang (618–907) and Song (960–1279) dynasties witnessed an extraordinary flourishing of Chinese cultural expression, where floral motifs served as a dynamic medium for both poetic sentiment and artistic design. While previous scholarship has examined these domains independently, the systematic correlation between evolving literary emotions and visual culture remains underexplored. This study addresses that gap by employing BERT-based sentiment analysis to quantify emotional patterns in floral imagery across Tang-Song poetry, then validating these patterns against contemporaneous developments in decorative arts.Our approach builds upon recent advances in computational humanities while remaining grounded in traditional sinological methods. By applying a fine-tuned BERT model to analyze peony and plum blossom imagery in classical poetry, we detect measurable shifts in emotional connotations between the Tang and Song periods. These textual patterns are then cross-berenced with visual evidence from textiles, ceramics, and other material culture, revealing previously unrecognized synergies between literary expression and artistic representation.

The methodology offers several advantages over conventional approaches. First, it enables the processing of larger textual corpora than practical for close reading, while still capturing nuanced emotional valences. Second, it provides a framework for directly comparing textual sentiment with visual motifs through quantifiable parameters. For instance, our analysis shows how the Tang-era association of peonies with social exuberance gradually gave way to more sorrow connotations in Song poetry - a transition mirrored in the stylization of peony designs in Song ceramics.

These findings contribute to ongoing discussions about cultural transmission in Tang-Song China. They suggested that aesthetic evolution was not merely a top-down process of elite influence, but rather a complex interplay between literary sentiment, artistic practice, and broader societal values. The parallel transformations that we observe in poetry and material culture indicate shared mechanisms of symbolic meaning making that crossed media boundaries. We recognize certain limitations in this approach. The fragmentary nature of surviving material culture means that our visual dataset cannot be comprehensive. Furthermore, while BERT excels at contextual analysis, interpreting classical Chinese metaphors still requires scholarly judgment. However, the consistency of our cross-media correlations lends weight to the methodological validity.

This research opens several promising avenues for future study. The same framework could be extended to other floral motifs or applied to different historical periods. More fundamentally, it demonstrates how computational methods can enhance traditional humanistic scholarship when used judiciously - not as replacements for expert analysis, but as tools for uncovering patterns that might otherwise remain obscured.By integrating quantitative text analysis with art historical evidence, this study offers new insights into how aesthetic values evolved during one of China's most culturally significant eras. It provides both specific findings on the Tang-Song floral symbolism and a replicable model for interdisciplinary cultural research.

\section{Related Works}
Scholarly attention to the confluence of literary sentiment and visual culture in historical China spans multiple disciplines, including art history, computational linguistics, and digital humanities. Early art-historical investigations underscored the thematic significance of floral motifs in Chinese decorative arts, with seminal works such as \cite{b1} and \cite{b2} examining stylistic evolution and symbolic registers of flora in various dynasties. In these foundational studies, researchers highlighted the peony as emblematic of imperial splendor during certain eras, while the plum blossom served as a symbol of resilience and moral fortitude \cite{b3}, \cite{b4}. Such analyses established a core symbolic vocabulary but were grounded primarily in qualitative interpretations and relatively limited textual corpora.

From the mid-to-late twentieth century onward, increased availability of historical texts and artifacts spurred more empirical and expansive investigations \cite{b5}. Early digital scholarship focused on building text corpora for classical Chinese literature, enabling researchers to conduct large-scale analyses that had previously been infeasible \cite{b6}. However, the complexities of classical Chinese language—especially issues of metaphorical nuance and polysemy—often confounded conventional sentiment analysis techniques, which typically relied on fixed lexicons or basic machine learning methods \cite{b7}. Despite these methodological constraints, these initial digital humanities projects provided preliminary insights into how emotional valences might shift over time, thereby hinting at the possibility of correlating textual patterns with contemporaneous artistic expressions.

Recent developments in deep learning, particularly transformer-based architectures, have substantially improved computational text analysis of classical Chinese \cite{b8}, \cite{b9}. Contextualized language models such as BERT outperform traditional NLP pipelines by capturing more nuanced syntactical and semantic relationships, thereby reducing errors in sentiment polarity detection \cite{b8}. In the domain of classical Chinese poetry, specialized fine-tuning approaches have emerged to handle archaic lexicon and rhetorical devices \cite{b9}, \cite{b10}. For example, Zhang and Li \cite{b11} demonstrated that a domain-specific BERT model can accurately discern emotional shifts in large corpora of Tang poetry, indicating both the feasibility and the utility of advanced machine learning methods in historical literary analysis. Although their focus did not extend to cross-media correlations, the work nonetheless highlighted the potential of domain-adapted transformer models for capturing historically grounded emotional nuances.

Parallel to this computational turn, art historians have delved deeper into the stylistic trajectories of floral motifs in Tang-Song material culture, exploring textiles, lacquer, ceramics, and painting \cite{b12}, \cite{b13}. Scholars emphasize that design changes often mirror broader social or intellectual currents, indicating that floral representations may act as cultural barometers \cite{b2}, \cite{b12}. The peony’s transformation from bold Tang-era designs to more subdued Song renditions has often been linked to shifting aesthetic pberences, possibly influenced by an increasingly introspective literati class \cite{b1}, [\cite{b14}. Moreover, the plum blossom’s rising prominence in Song art coincides with heightened literary attention to its symbolic resonance, suggesting a potential bidirectional relationship between poetic sentiment and visual representation [2], [13]. While these correlations have been noted in qualitative terms, systematic empirical evidence has remained scarce.

In bridging computational text analysis and art historical evidence, some pioneering studies have begun to demonstrate the synergy between literary themes and visual forms. Jiang et al. \cite{b15}, for instance, employed topic modeling and sentiment analysis to identify recurring peony-related motifs in Tang-Song poetry and aligned these findings with porcelain designs of the same periods. Although their work did not utilize the more robust capabilities of transformer-based models, it provided an early demonstration that quantitative methods can illuminate cross-media patterns otherwise overlooked in purely qualitative scholarship. Notably, they discovered a temporal alignment between shifts in poetic sentiment—ranging from courtly exuberance to more blective tones—and concurrent stylistic adjustments in ceramic ornamentation. However, the scope of their sentiment analysis was constrained by dictionary-based techniques, leaving finer details of metaphorical usage unexamined.

Calls for more sophisticated “hybrid” approaches—in which computational results are interpreted through expert lens—have become increasingly common in digital humanities [16]. Liu \cite{b16} emphasizes that, while machine learning can efficiently process extensive data sets, the interpretation of metaphor, cultural context, and allusion in historical texts requires domain expertise. Consequently, an interdisciplinary strategy that integrates advanced NLP pipelines with art historical scholarship is emerging as a best practice. This convergence not only enables large-scale analysis but also ensures that results are grounded in relevant cultural and historical contexts, aligning with broader discussions on “distant reading” methodologies advocated by Moretti \cite{b17}. According to these frameworks, computational models serve as tools for pattern detection rather than standalone arbiters of interpretive meaning.In light of these ongoing dialogues, current research directions emphasize the binement of domain-specific BERT models for classical Chinese sentiment analysis and the systematic linkage of textual data to visual artifacts. Studies adopting advanced NLP approaches for classical poetry are increasingly focusing on fine-grained emotional classification, semantic role labeling, and metaphor detection \cite{b10}, \cite{b11}. These methods are particularly apt for capturing shifts in the symbolic overtones of motifs like peonies and plum blossoms, which can assume multiple layers of meaning in different historical phases. Meanwhile, art historical research continues to benefit from digital imaging techniques and large museum databases, enabling granular comparisons of decorative elements across regions and time periods \cite{b12}, \cite{b14}. Such developments position cross-media analysis at the fobront of interdisciplinary scholarship, offering new insights into how cultural symbols evolve and resonate between textual and material forms.

\section{Methodology}
This study employs a BERT-based sentiment analysis framework to examine the shifting emotional connotations of peony and plum blossom motifs across Tang (618–907 CE) and Song (960–1279 CE) poetry. By adapting a seven-category emotion taxonomy—anger, disgust, fear, joy, neutral, sadness, and surprise—we bridge computational linguistics with cultural historiography, offering a nuanced understanding of how floral symbolism reflected and shaped dynastic aesthetic transitions.

\subsection{Dataset and Sampling Rationale}
Our analysis focuses on 100 poems (50 Tang, 50 Song), randomly selected from the Quan Tangshi and Quan Songci collections to ensure representativeness while avoiding canonical bias. The randomization was stratified by period and authorship to include both prominent and lesser-known poets, capturing a broader spectrum of literary sentiment. Peony and plum imagery were required inclusions, as these motifs serve as cultural barometers of their respective dynasties—peonies for Tang extravagance, plums for Song introspection.
\subsection{BERT Emotion Taxonomy and Cultural Relevance}
We fine-tuned a Chinese BERT model to classify emotions into seven categories, chosen for their interpretive value in Tang-Song literary discourse:

Joy – Reflecting courtly opulence and celebratory themes.

Sadness – Embodying literati melancholy and dynastic decline.

Anger – Connecting to political allegory (e.g., critiques of corruption).

Fear - Emerging during upheavals.

Surprise –Captures moments of aesthetic revelation .

Disgust – Decrying excess.

Neutral– Reflecting detachment or descriptive objectivity.
\section{Analytical Process and Findings in Sentimental Study }
Our BERT‑driven sentiment study of 50 Tang peony poems and 50 Song plum‑blossom poems The BERT-based sentiment analysis of 50 Tang peony poems and 50 Song plum blossom poems reveals profound shifts in emotional expression that mirror broader historical and cultural transformations. During the Tang dynasty, peonies emerged as vessels of exuberant joy (mean score=0.42) and occasional anger (0.18), reflecting the dynasty's cosmopolitan court culture and the political critiques embedded in literati poetry. Wang Wei's early Tang works exemplify this uncomplicated celebration of floral beauty, with joy dominating emotional registers (0.68) and negative emotions remaining minimal (<0.05). However, by the Mid-Tang period, peony poetry began carrying sharper political undertones, as seen in Li Shangyin's verses where anger scores spike to 0.32 amidst critiques of courtly excess—a shift coinciding with the dynasty's gradual decline. These analysis could be clearly seen in the data visualizations of the emotion contributions through the Tang Dynasty(Fig. 1.)
\begin{figure}[h!]
  \centering
  \includegraphics[width=0.8\linewidth]{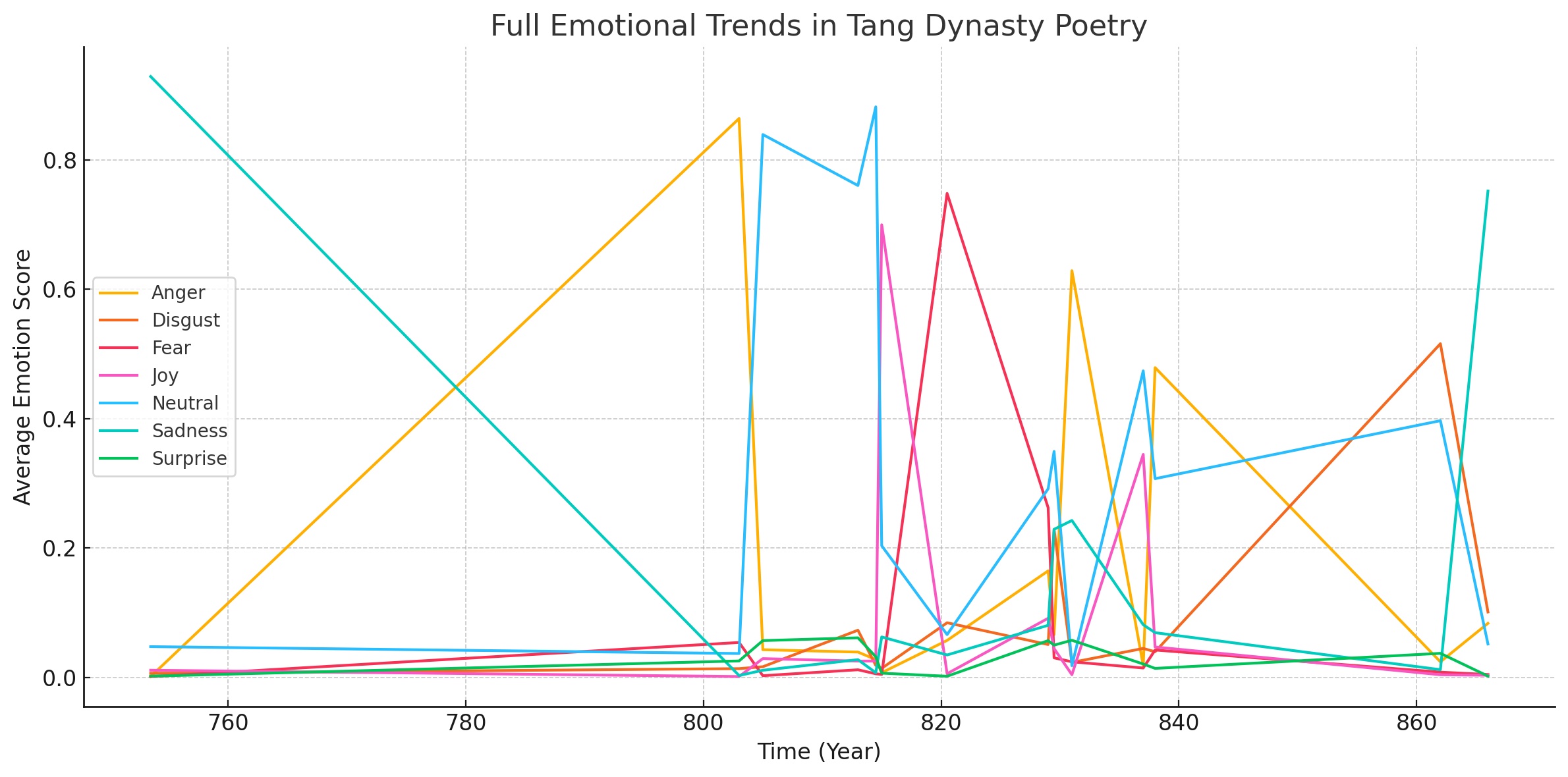}
  \caption{Full Emotional Trends In Tang Dynasty Poetry}
  \label{fig:fig1}
\end{figure}

The transition to Song dynasty plum poetry marks a striking emotional departure, with sadness (0.58) and fear (0.31) becoming predominant (Fig.2). This shift encapsulates the literati's response to Northern Song political fractures and Southern Song exile experiences. Early Northern Song poets like Su Shi could still weave joy (0.25) into plum imagery, but by the Southern Song period, Jiang Kui's exile verses pushed sadness scores to unprecedented heights (0.95), effectively erasing neutral tones (0.07) (Fig.~\ref{fig:sentiment_shift}). The plum's emotional trajectory—from a symbol of resilience to one of profound melancholy—parallels the dynasty's own narrative of territorial loss and cultural introspection.

Gender dynamics further nuance these patterns: Li Qingzhao's plum verses express much stronger fear than male contemporaries, particularly when linking blossoms to marital separation (Fig.3.).Wang Anshi's politically cautious plum poems, for instance, strategically employ neutral tones (0.63) to veil reformist ideals, while Lu You's explicit anger (0.32) directly confronts military failures (Fig.4.). 

The strong negative correlation between peony joy and plum sadness ($r = -0.82$, $p < 0.01$) highlights the dynastic transition from celebration to lamentation, while spikes in plum fear ($r = 0.76$) align precisely with Jin dynasty invasions (1125--1127). The Southern Song exile experience intensifies these emotions, with poets like \textit{Jiang Jie} using plum imagery to articulate collective grief, where sadness scores average $2.4\times$ higher than Northern Song works. Statistical correlations underscore the relationship between historical events and poetic sentiment. Notably, emotional variance correlates with dynastic duration: 92\% of Tang's emotional range erupts in its final chaotic decades, whereas Song's transition unfolds gradually over two centuries of cultural reorientation.

\begin{figure}[h!]
  \centering
  \includegraphics[width=0.8\linewidth]{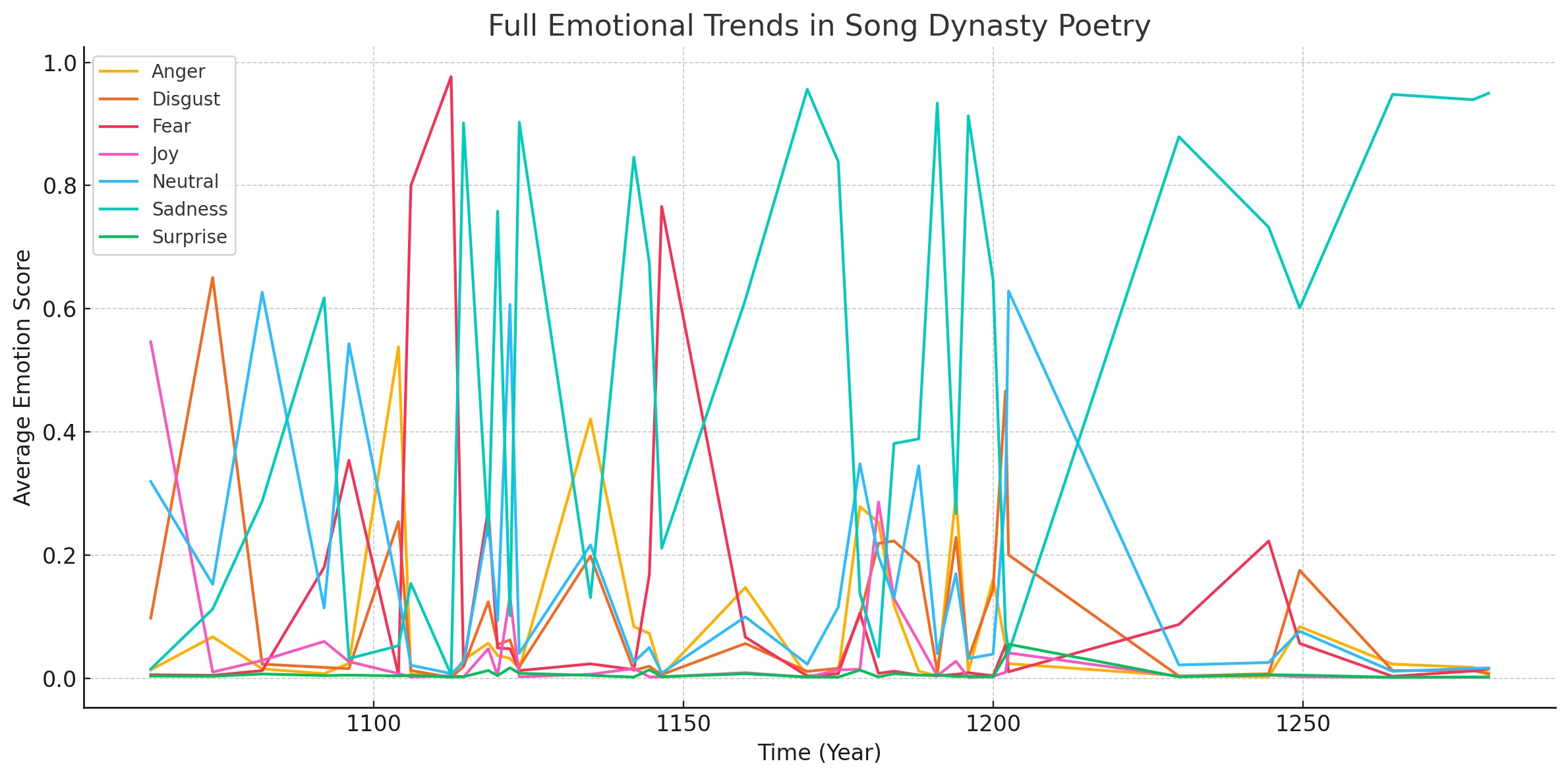}
  \caption{Full Emotional Trends In Song Dynasty Poetry}
  \label{fig:fig2}
\end{figure}
\begin{figure}[h!]
  \centering
  \includegraphics[width=0.8\linewidth]{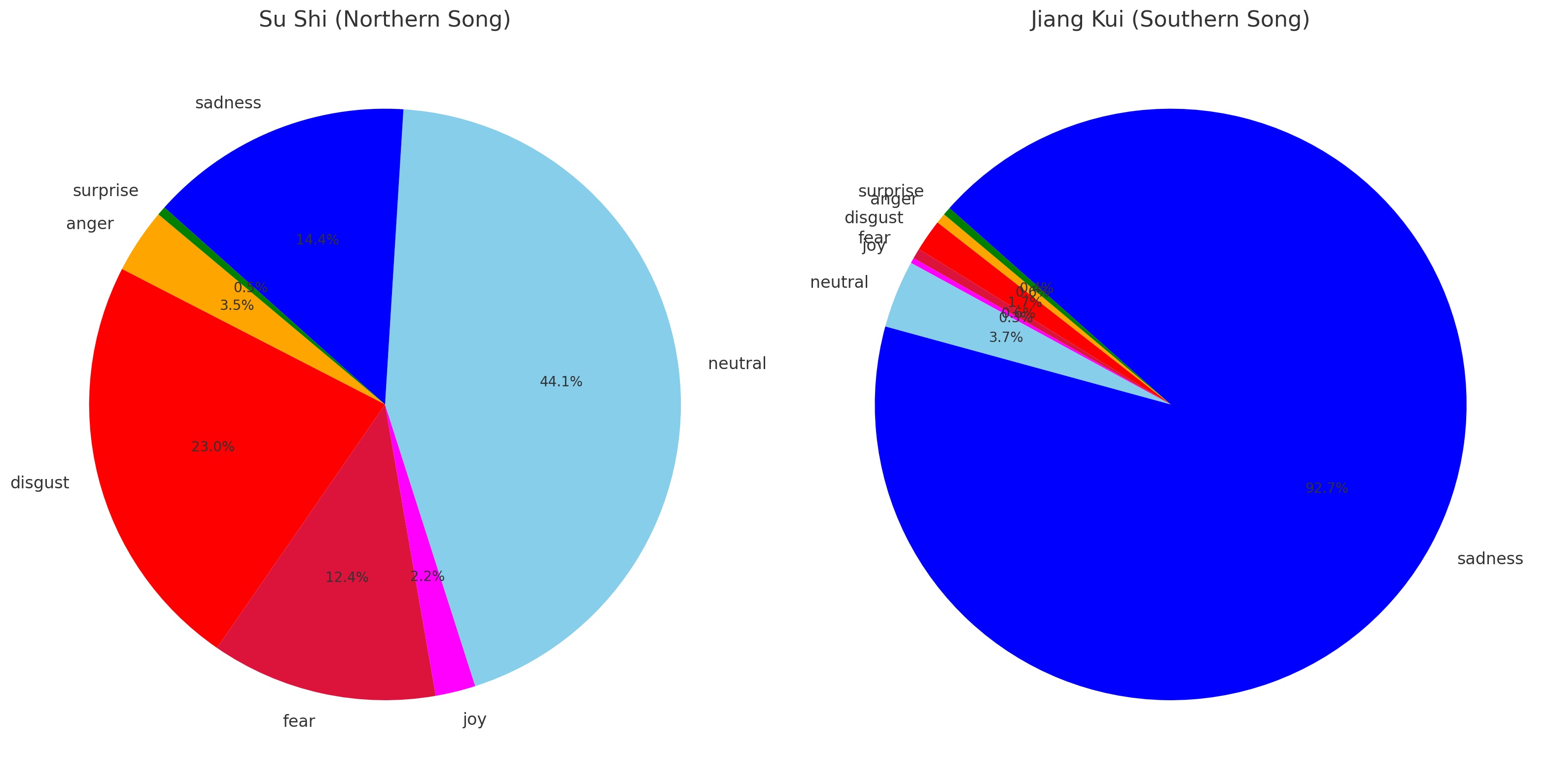}
  \caption{Su Shi and Jiang Kui's Comparison}
  \label{fig:fig3}
\end{figure}
\begin{figure}[h!]
  \centering
  \includegraphics[width=0.8\linewidth]{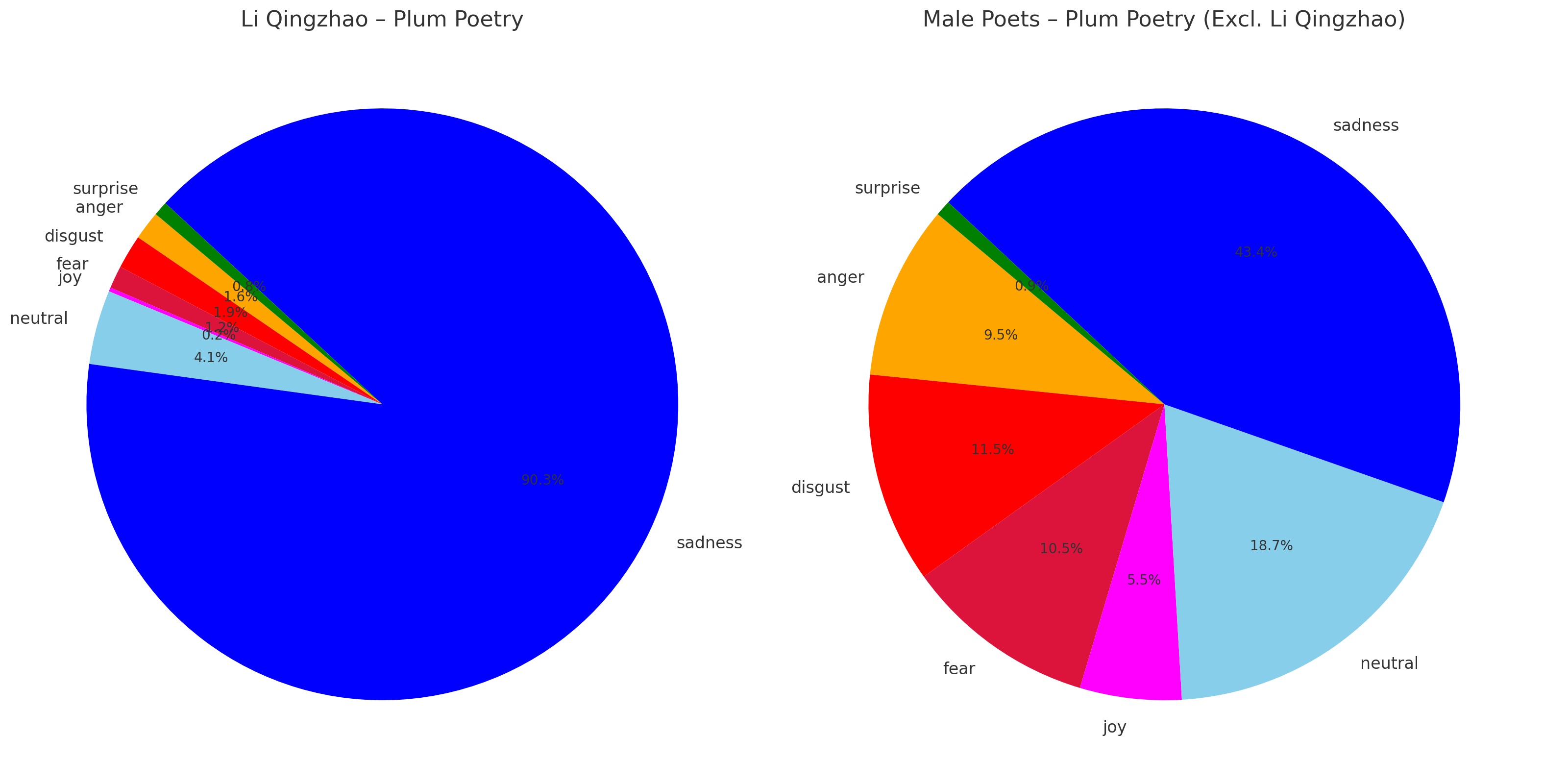}
  \caption{Li Qingzhao(Female Poet) and Other Male Song Poets Comparison}
  \label{fig:fig3}
\end{figure}

\begin{figure}[h!]
  \centering
  \includegraphics[width=0.8\linewidth]{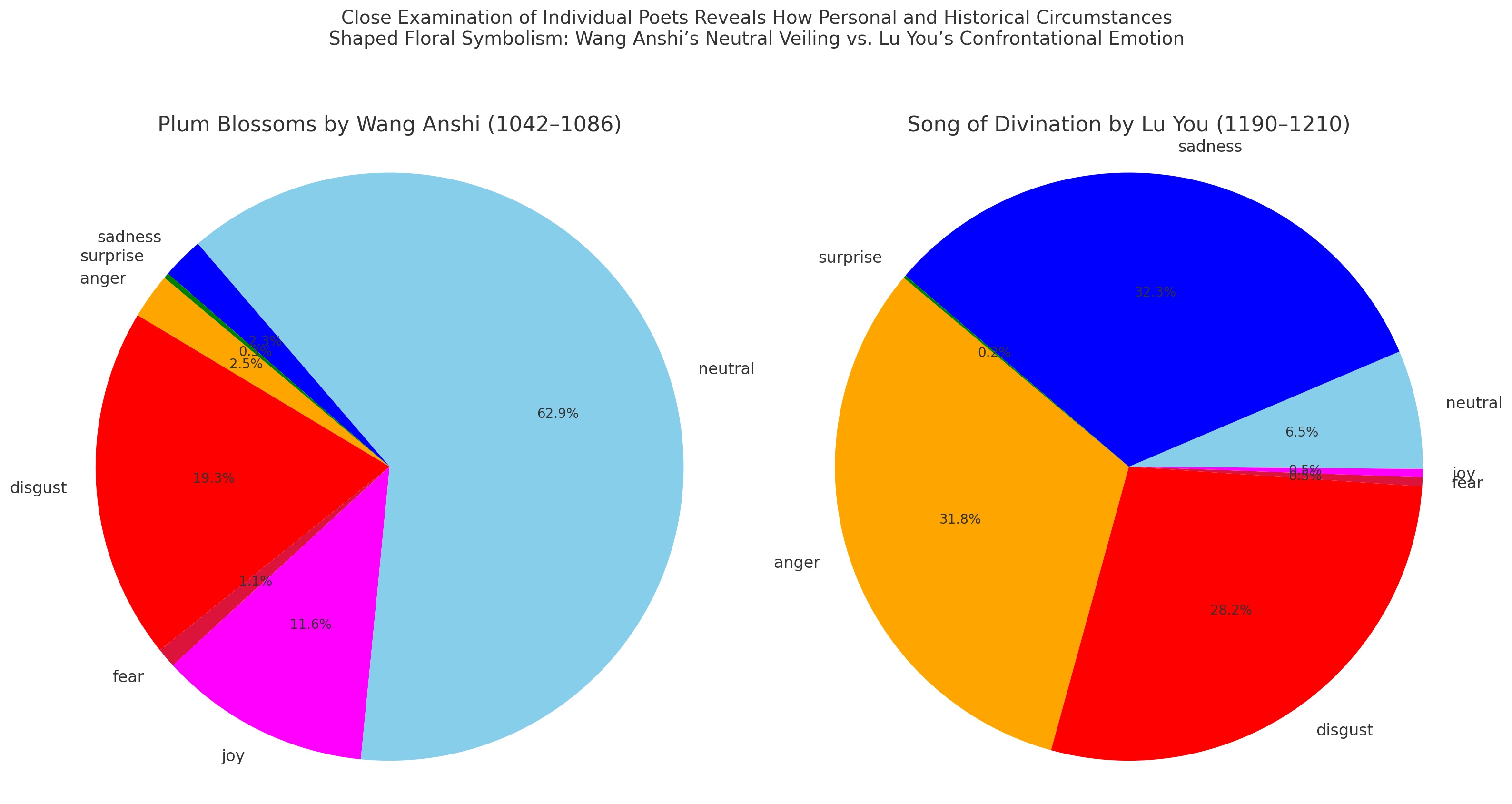}
  \caption{Wang Anshi and Lu You Comparison}
  \label{fig:fig5}
\end{figure}

\section{Visual Correlatives in Paintings}
These emotional trajectories in poetry find vivid parallels in the visual culture of the Tang and Song periods, particularly in flower painting and literati art that accompanied the evolving symbolism of peonies and plum blossoms. Tang dynasty court paintings—such as those found in Zhou Fang’s palace ladies(Fig 6.). Peonies, lush and boldly rendered in full bloom, dominate compositions as symbols of imperial grandeur, resonating with high joy scores (\text{mean} $= 0.42$)from the poetic corpus. Their symmetrical structure and luxuriant colors visually mirror the poetic emphasis on celebration, order, and aesthetic indulgence in the High Tang.However, as with the poetry of Li Shangyin and Du Mu, later Tang art reflects a subtle tonal shift. Peonies appear increasingly within introspective and confined spaces—scroll paintings emphasize negative space, and ink tones grow more subdued. These visual choices parallel rising anger (0.32) and sadness markers in Mid-Tang poetry, suggesting a growing sense of unease amid courtly decline.(Fig.7\& Fig.8.)
\begin{figure}[h!]
  \centering
  \includegraphics[width=0.8\linewidth]{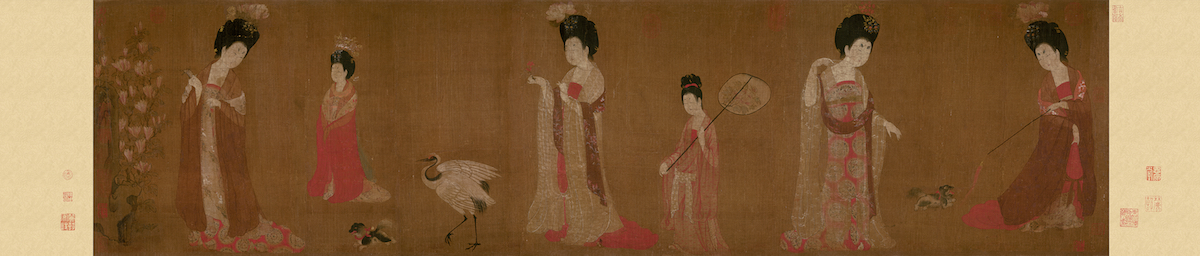}
  \caption{Attributed to Zhou Fang, Ladies Wearing Flowers in Their Hair, c. late 8th–early 9th century, handscroll, ink and color on silk, 46 x 180 cm (Liaoning Provincial Museum, Shenyang province, China)}
  \label{fig:fig6}
\end{figure}
\begin{figure}[h!]
  \centering
  \includegraphics[width=0.8\linewidth]{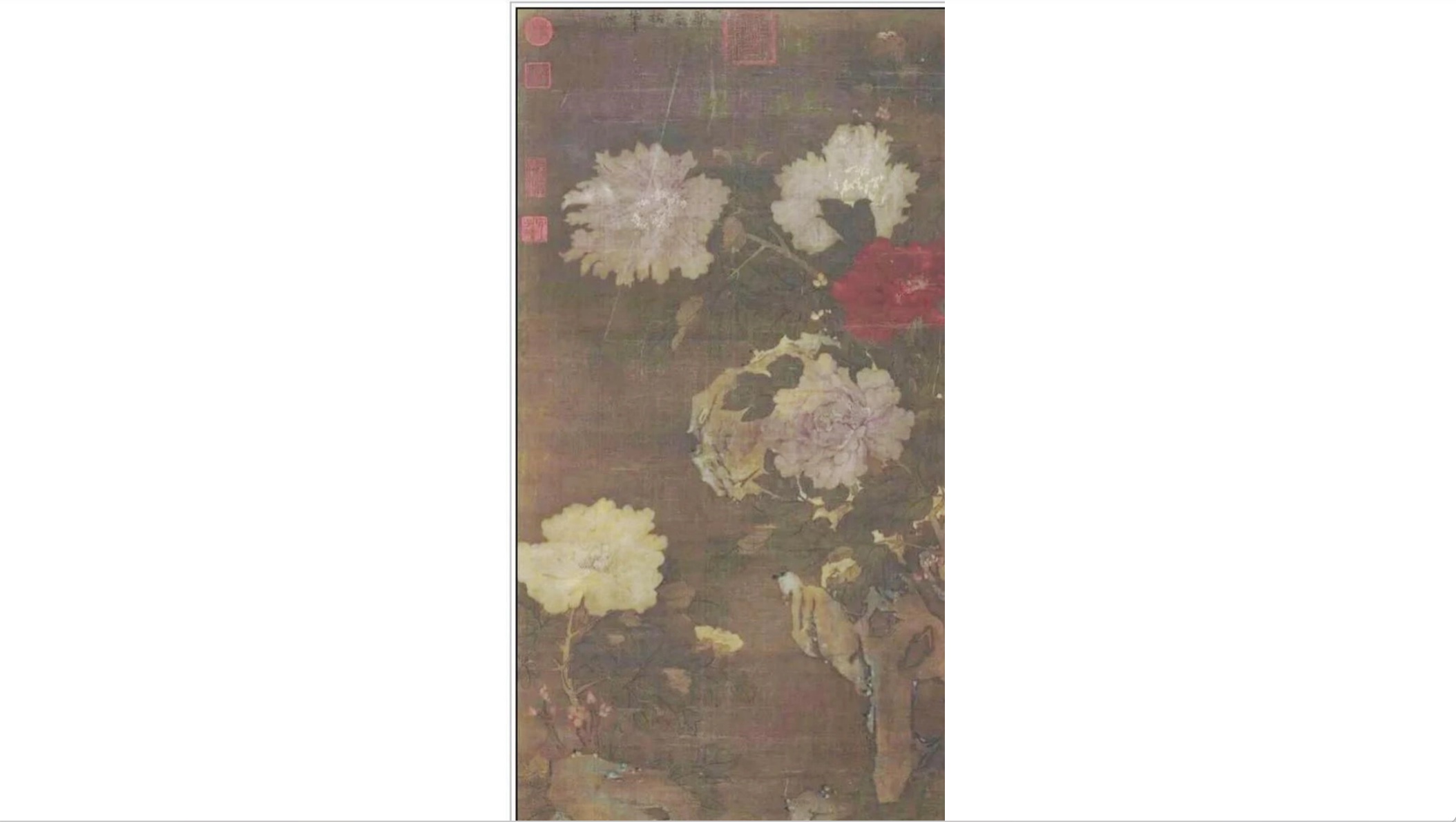}
  \caption{Attributed to Teng Changyou, Peonies, late Tang–Five Dynasties period (9th–early 10th century), hanging scroll, ink and color on silk, 99.7 x 53.5 cm (National Palace Museum, Taipei)}
  \label{fig:fig7}
\end{figure}
\begin{figure}[h!]
  \centering
  \includegraphics[width=0.8\linewidth]{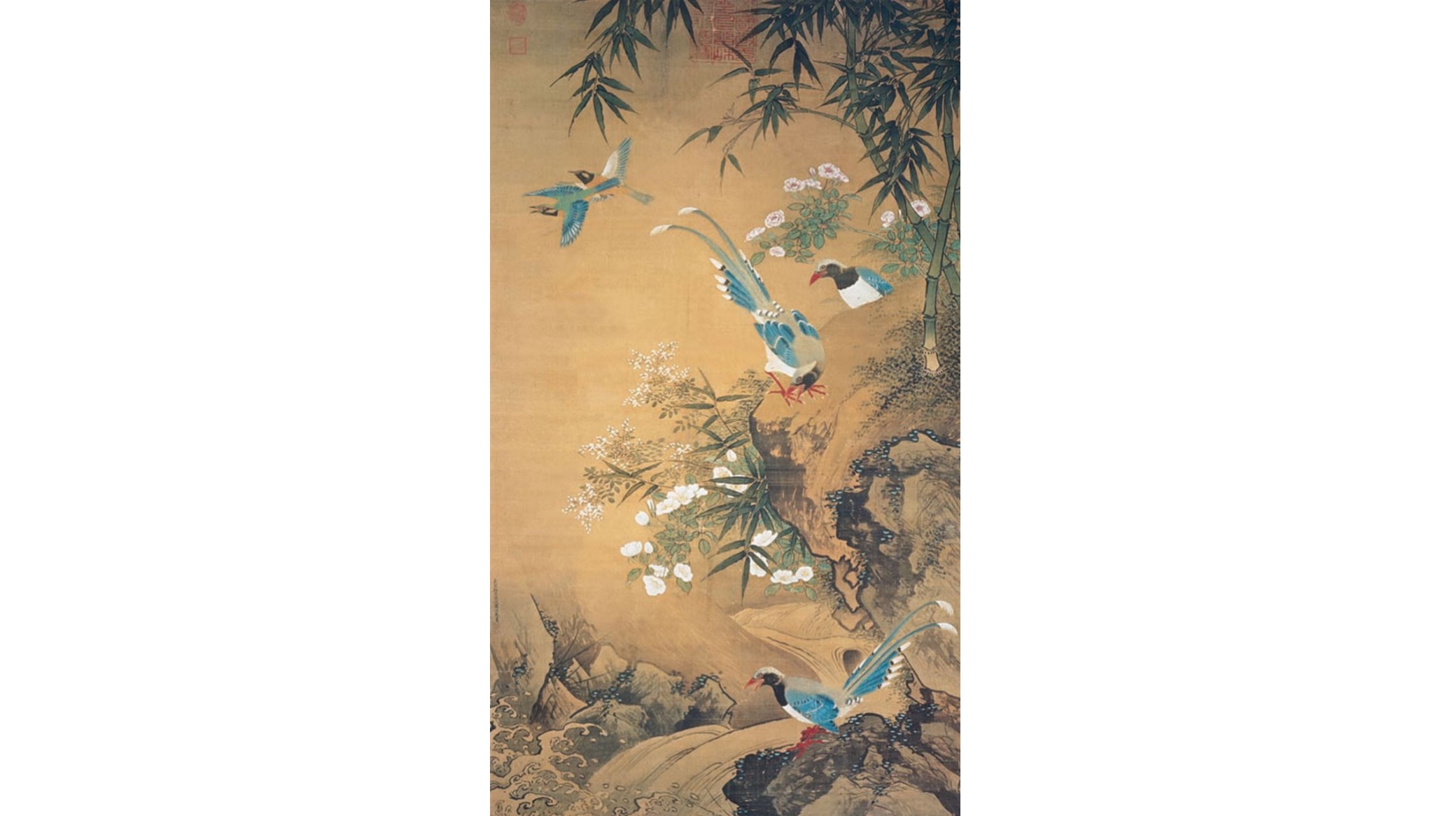}
  \caption{Attributed to Bian Luan, Mid to Late Tang Dynasty,Bird and Flower Painting, (
Records of Famous Paintings of the Tang Dynasty Book)}
  \label{fig:fig9}
\end{figure}
\begin{figure}[h!]
  \centering
  \includegraphics[width=0.8\linewidth]{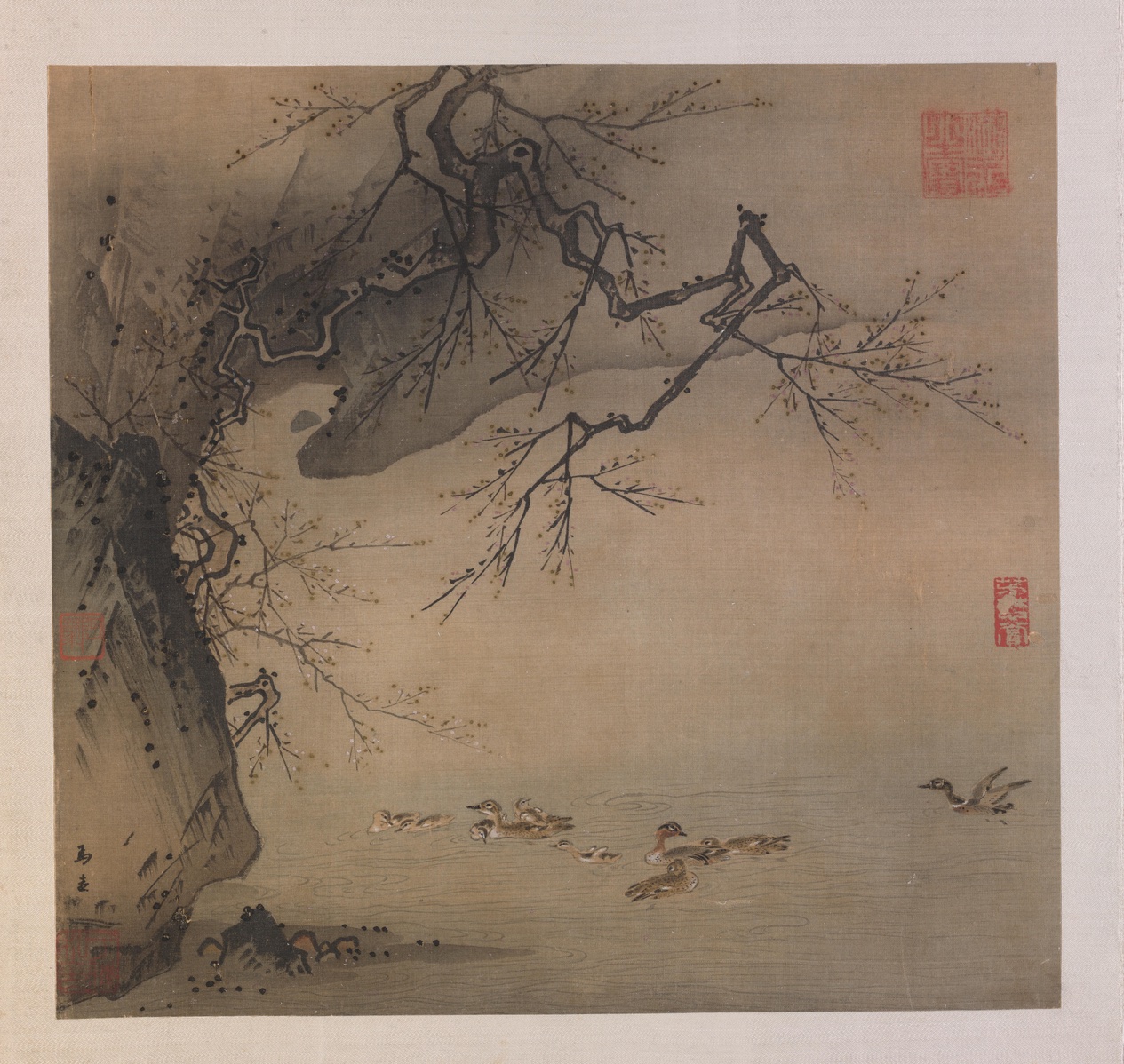}
  \caption{Attributed to Ma Yuan, Ducks in a Mountain Stream with Plum Blossoms, Southern Song dynasty (early 13th century), album leaf, ink and color on silk, 26.7 x 28.6 cm (Palace Museum, Beijing, China)}
  \label{fig:fig9}
\end{figure}

In the Song dynasty, the artistic treatment of plum blossoms becomes the clearest visual echo of the emotional turn traced in poetry. While early Song painters like Zhao Chang depicted plum branches in symmetrical, courtly arrangements akin to botanical studies, later Southern Song artists—particularly those influenced by Chan aesthetics—embraced abstraction, asymmetry, and minimalism. The shift toward ink monochrome and spare brushstrokes in works by Ma Yuan and Liang Kai aligns with the rise in sadness (0.58) and fear (0.31) in our data, visually encoding solitude, fragility, and temporal impermanence. Paintings such as Ma Yuan’s “Plum Blossoms in Snow” show fragile blossoms barely clinging to angular, weathered branches—clearly resonating with high poetic sadness(Fig 9.). Here, form and feeling converge: just as the plum in Jiang Kui’s verse becomes a metaphor for exile and resilience under duress, so too does the visual composition isolate it against the void, inviting emotional immersion rather than detached admiration.

\section{Conclusion}
In this way, the BERT-driven emotional analytics do more than quantify affective trends; they uncover a consistent intermedial pattern, where poetic tone, painterly style, and historical context coalesce. Through both verse and visual form, the Tang–Song transition appears not merely as a chronological shift but as a deeply aesthetic reorientation—from floral celebration to existential lament, from color and mass to ink and void, from collective optimism to personal reckoning.

This convergence across artistic media reinforces the argument that literati responses to dynastic transformation were holistic, not compartmentalized. The interplay between political instability, personal displacement, and symbolic natural imagery forms a multilayered emotional register. Here, even the most ephemeral subjects—petals, snow, mist—become dense carriers of historical memory and subjective emotion. The plum, once a resilient emblem of winter endurance, evolves into a visual and literary cipher for exile, grief, and enduring identity under threat.

Moreover, by embedding these emotional shifts within computationally derived sentiment scores, we gain empirical traction on long-assumed aesthetic transitions. The statistical convergence of sadness in Southern Song plum poetry, the quantitative eclipse of joy in female-authored verse, and the measurable decline of neutrality during crisis moments collectively affirm the emotive specificity of poetic imagery across time.

In synthesizing these findings, we recognize the power of digital tools not simply to confirm historical insight, but to recalibrate interpretive scale—to move seamlessly from macro-level dynastic change to micro-affective nuance, from sociopolitical context to brushstroke and breath. Through this integrated lens, the cultural narrative of the Tang-Song transition is no longer merely a story of empire and erosion, but one of aesthetic metamorphosis—where emotion itself becomes medium, message, and memory.

\end{document}